\documentclass[a4paper,11pt]{article}

\usepackage{amsmath,amssymb,amsfonts}
\usepackage{algorithmic}

\usepackage{textcomp}
\usepackage{xcolor}
\usepackage{todonotes}
\usepackage{booktabs}
\usepackage{multirow}

\usepackage{fourier}
\usepackage{color}
 \usepackage{graphicx}
\usepackage{url}
\usepackage[affil-it]{authblk}
\usepackage{wrapfig}

\usepackage[T1]{fontenc}
\usepackage{times}

\begin{document}

\title{Defect Classification in Additive Manufacturing
Using
CNN-Based Vision Processing}

\author{Xiao Liu, Alessandra Mileo and Alan F. Smeaton}
\affil{Dublin City University, Glasnevin, Dublin 9, Ireland\\
alan.smeaton@dcu.ie}
\date{}
\maketitle
\thispagestyle{empty}

\begin{abstract}

The development of computer vision and in-situ monitoring using visual sensors  allows the collection of large datasets from the additive manufacturing (AM) process. Such datasets could be used with machine learning techniques to improve the quality of AM. This paper examines two scenarios: first, using convolutional neural networks (CNNs) to accurately classify defects in an image dataset from AM and second, applying active learning techniques to the developed classification model. This allows the  construction of a human-in-the-loop mechanism to reduce the size of the data required  to train and generate training data.    

\end{abstract}

%%%%%%%%%%%%%%%%%%%%%%
\section{Introduction}
Large and openly available datasets of annotated images containing up to millions of training examples such as  Pascal VOC \cite{everingham2010pascal} are available to machine learning researchers for many applications. This has enabled huge improvements in machine learning over recent years. By contrast, such openly available datasets are not available  in the domain of Additive Manufacturing (AM) or 3D printing because labelled samples are difficult, expensive, and time-consuming to obtain as shown in \cite{QIN2022102691} and \cite{WANG2020101538}. As a result of poor data availability, 
researches in  AM  often have to use only a limited amount of labelled samples for  training tasks
before then leveraging a large number of unlabelled image data. Some researchers have called this  the ``small data challenge in the big data era” \cite{qi2020small}. 

To overcome this challenge, we present a method that applies transfer learning and fine-tuning on a CNN-based neural network model to achieve accurate classification of manufacturing defects. This uses a dataset of images of the melt pool, created 
from the interaction between a laser and the materials used in  manufacturing, taken during the AM process.  Structural defects in the resulting  output can sometimes be detected during manufacture  from observations of the melt pool.
Our technique involves using active learning algorithms to  reduce the number of labelled samples required in the training process. We perform automatic labelling using the model to generate larger datasets of labelled images from unlabelled samples, for use in training.

\section{Methods}
%The next section presents the methodology of our approach.
{\it Transfer learning} is a method that performs training a neural network model using data from a source domain then later applying the trained model to a target domain that is different from the source. This  allows rapid progress in re-training   and significantly reduces the required number of training samples in the target domain. It is commonly used in computer vision tasks  such as classification  to support improved performance in domains which are data-poor. 
In recent years, transfer learning has proved to be effective in the task of defect classification in AM, such as the work presented in \cite{liu2021deep} and \cite{WESTPHAL2021101965} where transfer learning and fine-tuning were applied to the training of CNN based neural network architectures. 

{\it Active learning} \cite{settles.tr09} is a technique for labelling data that selects and prioritises the most informative data points to submit to an annotator for labelling. Such prioritised data points have the highest potential impact on the supervised training of a machine learning model, thus accelerating the training process. The combination of transfer learning and active learning allows leveraging small amounts of labelled data to improve the performance of the training process of a deep learning model. 

% I will give more detailed illustrations in the next section.

\section{Classification Experiments in Additive Manufacturing}
\label{sec:exp}
To investigate the potential for transfer learning and active learning in the task of defect detection in AM, a case study was carried out using the open image dataset from \cite{WESTPHAL2021101965}. This contains 4,000 images, manually divided into 2 different  defect detection classes in AM. The images in this dataset are clearly separated into 3 balanced subsets for training (2,000), testing (1,000) and validation (1,000).

To conduct experiments, we employed a VGG16 based classifier from  previous work which proved to be accurate in the task of defect classification on  images generated from emission monitoring during additive manufacturing  \cite{anon2022}. This classifier relies on transfer learning in which  13 convolutional layers from a pre-trained VGG16 model are used for feature extraction and the weights in these layers had been  trained using ImageNet data. After the convolutional layers, 2 dense layers with ReLU activation function are added  followed by 1 dense layer as the output layer using Sigmoid as the activation function, since the targeted dataset are divided into 2 classes for binary classification. In the original paper  \cite{WESTPHAL2021101965}, the best classification performance is generated  using a VGG16 based CNN model which is the reason we do not use a more recent model such as ResNet. We consider that as a baseline  for further investigation in this study. 

The tuning of hyperparameters involves adjusting the optimiser, learning rate, batch size and training epochs. There are 3 optimisers in the test we use which are Adam, SGD and RMSprop in combination with learning rate in a range from 10$^{-2}$ to 10$^{-5}$. We have also conducted training using different batch sizes (4, 32, 64) and training epochs (30, 60, 120). The cost function used in all tests is binary cross entropy. To reduce overfitting, weight regularisers are added to the 2 dense layers with the ReLU activation function mentioned above. The weight decay regulariser, also known as L2 regulariser which calculates the sum of the squared weights, is applied when initialising the keras model. The tuning of this hyperparameter is in a range from 10$^{-1}$ to 10$^{-4}$ and tested for multiple times until no obvious overfitting issue appears in the training and validation. 

After tuning on hyperparameters for multiple combinations, the best preforming combinations regarding the 3 types of optimisers are shown in Table~\ref{tab:baseline_test} together with  classification results on the validation dataset in comparison with the baseline from \cite{WESTPHAL2021101965}. These initial tests were performed to check how adaptive our approach is on this dataset. The results show that all  3 optimisers can reach a value around 98\% of the validation accuracy and our classification model is well-adapted to this dataset. The results also show that for this dataset a smaller batch size used in the training process such as 4, gives better performance and this can be explained as smaller batch sizes require more frequent weight updates during training. In turn this can help the model adjust its parameters more quickly and respond to changes in the data distribution which increases the model's ability to adapt to a new dataset. Finally, although not shown here, accuracy is stable thoughout the training  showing no overfitting.
%==================Table 1==================
%i'm creating a Bigger table with all the results
% Please add the following required packages to your document preamble:
% \usepackage{booktabs}
% \usepackage{multirow}
\begin{table}[htbp]
\centering
\caption{Best performing hyperparameters for each  optimiser, performance results on the validation set. Results marked `*' are updates provided directly to us by the authors of \cite{WESTPHAL2021101965} in response to us pointing out errors in the original paper. An author correction to the  copy of record is now underway.}
\label{tab:baseline_test}
\resizebox{0.9\columnwidth}{!}{%
\begin{tabular}{@{}lccccccccc@{}}
\toprule
\multicolumn{1}{c}{\begin{tabular}[c]{@{}c@{}}Experiment: \\ Optimiser, learning rate\end{tabular}} &
  Batch &
  Epochs &
  \multicolumn{2}{c}{\begin{tabular}[c]{@{}c@{}}Confusion \\ matrix\end{tabular}} &
  Accuracy &
  Precision &
  Recall &
  F1-Score &
  AUC \\ \midrule
\multirow{2}{*}{Baseline} &
  \multirow{2}{*}{64} &
  \multicolumn{1}{c|}{\multirow{2}{*}{30}} &
  \multicolumn{1}{c|}{496} &
  \multicolumn{1}{c|}{4*} &
  \multirow{2}{*}{0.977*} &
  \multirow{2}{*}{0.992*} &
  \multirow{2}{*}{0.963*} &
  \multirow{2}{*}{0.977*} &
  \multirow{2}{*}{0.993} \\ \cmidrule(lr){4-5}
 &
   &
  \multicolumn{1}{c|}{} &
  \multicolumn{1}{c|}{19} &
  \multicolumn{1}{c|}{481} &
   &
   &
   &
   &
   \\ \midrule
\multirow{2}{*}{SGD, lr=0.01} &
  \multirow{2}{*}{4} &
  \multicolumn{1}{c|}{\multirow{2}{*}{60}} &
  \multicolumn{1}{c|}{483} &
  \multicolumn{1}{c|}{17} &
  \multirow{2}{*}{0.979} &
  \multirow{2}{*}{0.967} &
  \multirow{2}{*}{0.992} &
  \multirow{2}{*}{0.979} &
  \multirow{2}{*}{0.998} \\ \cmidrule(lr){4-5}
 &
   &
  \multicolumn{1}{c|}{} &
  \multicolumn{1}{c|}{4} &
  \multicolumn{1}{c|}{496} &
   &
   &
   &
   &
   \\ \midrule
\multirow{2}{*}{Adam, lr =0.00001} &
  \multirow{2}{*}{4} &
  \multicolumn{1}{c|}{\multirow{2}{*}{120}} &
  \multicolumn{1}{c|}{490} &
  \multicolumn{1}{c|}{10} &
  \multirow{2}{*}{0.988} &
  \multirow{2}{*}{0.980} &
  \multirow{2}{*}{0.996} &
  \multirow{2}{*}{0.988} &
  \multirow{2}{*}{0.998} \\ \cmidrule(lr){4-5}
 &
   &
  \multicolumn{1}{c|}{} &
  \multicolumn{1}{c|}{2} &
  \multicolumn{1}{c|}{498} &
   &
   &
   &
   &
   \\ \midrule
\multirow{2}{*}{RMSprop lr =0.00001} &
  \multirow{2}{*}{4} &
  \multicolumn{1}{c|}{\multirow{2}{*}{60}} &
  \multicolumn{1}{c|}{485} &
  \multicolumn{1}{c|}{15} &
  \multirow{2}{*}{0.982} &
  \multirow{2}{*}{0.971} &
  \multirow{2}{*}{0.994} &
  \multirow{2}{*}{0.982} &
  \multirow{2}{*}{0.997} \\ \cmidrule(lr){4-5}
 &
   &
  \multicolumn{1}{c|}{} &
  \multicolumn{1}{c|}{3} &
  \multicolumn{1}{c|}{497} &
   &
   &
   &
   &
   \\ \bottomrule
\end{tabular}
}
\end{table}

%===================END of Table 1=====================

\section{Active Learning Experiments in Additive Manufacturing}

Having developed a classifier which uses domain transfer across AM image datasets, we extended  training  to include  active learning  applied to further investigate classification performance during the progression of  AL iterations. 
The second experiment was performed in a series of  steps of (1) active sample section, (2) query for label, (3) train with queried sample, and (4) validate for current query iteration.   The cycle  iterates until a human supervisor  decides to complete the training phase when  validation accuracy achieves a target level.      

Here we apply a pool-based sampling scenario and an uncertainty sampling query strategy \cite{settles.tr09}. This is the most commonly used query strategy to start  generalised sampling  on this particular AM dataset.
The implementation of  active learning  uses Python 3 and Google Colab. During the experiment, a classifier model is  initialised and the optimser chosen is SGD as we found this gives more stabilised performance in the validation test and has minimal overfitting  even when  training is continued long after convergence. While Adam and RMSprop converge faster, there are larger fluctuations in the validation and minor overfitting after  training reaches convergence. In addition, though SGD yields a result  lower than the other 2 optimisers, it has slightly better potential that can be improved by applying active learning.   
During this experiment,  2,000 training samples were fed to the classifier with a  total of 40 queries and for each query 50 samples were actively selected by the uncertainty sampling query strategy. 

%It worth noting that, in this learning process, all the training samples are tread as not have been labelled as the uncertainty sampling query strategy calculate the uncertainty from prediction. 

The selected and queried samples were assigned a label by an annotator after which the newly labelled samples were used to fine-tune the classifier to improve   performance.  This was evaluated using classification accuracy on the validation dataset at the end of each query iteration and later we show results on the test set. 

\begin{figure}[htbp]
\centerline{\includegraphics[width=0.7\textwidth]{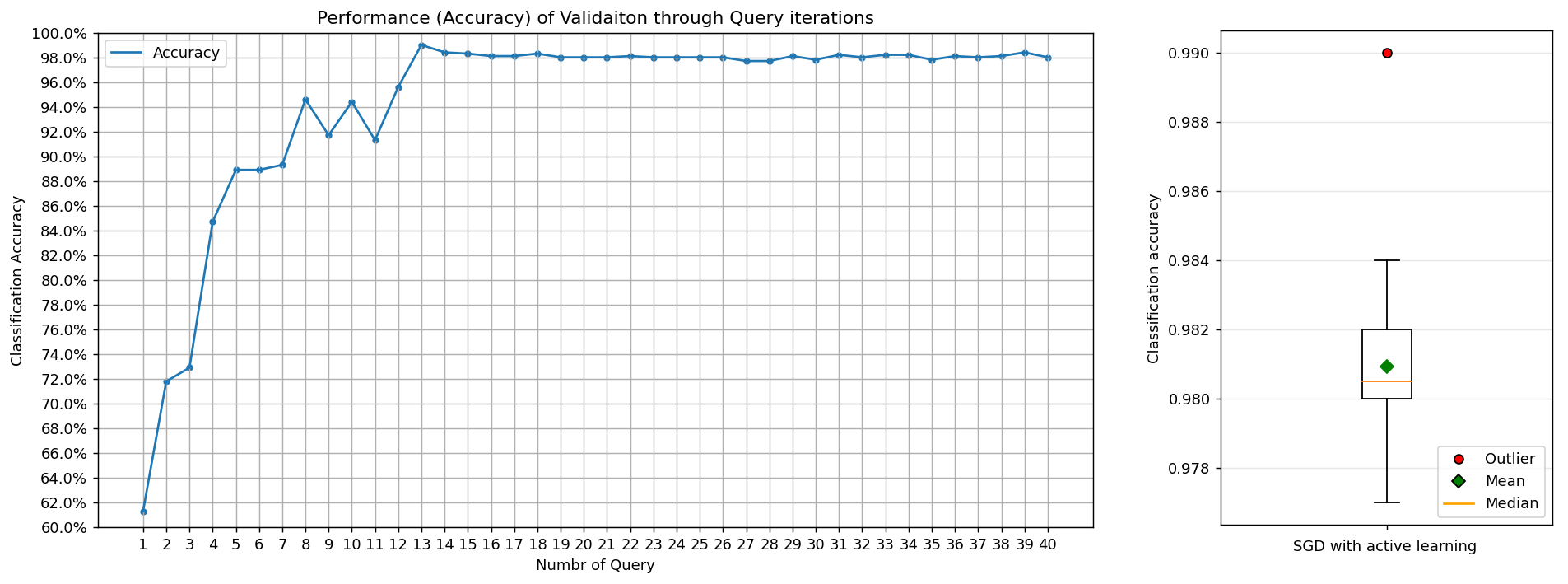}}
\caption{classification accuracy on the validation dataset at the end of each query iteration}
\label{fig:figAL_SGD}
\end{figure}

Following the inclusion of active learning,  validation accuracy in each query iteration is shown in Figure~\ref{fig:figAL_SGD} where results show that with the aid of active learning, the model reaches convergence after the 13th query and the value of validation accuracy is around 98\%. More specifically, the calculated mean value from the 13th to 40th queries is 0.981 with a standard deviation of 0.0246 and a peak  of 0.990. This  is slightly higher than the result of the SGD optimiser based model shown in Table~\ref{tab:baseline_test} and 1\% higher than the baseline. Overall performance after convergence is also relatively stable.  Results also show that  the model only needs the first 650 most informative samples to achieve the best performance which is only 37.5\% of the total 2,000 labelled training data.    
This trained model was used to classify the labels on the testing dataset mentioned in Section~\ref{sec:exp}, which is a balanced dataset consisting of 1,000 samples and the results are shown in Table~\ref{tab:test}. 
%text below are some 
%The results shows that the overall accuracy is 0.984 in the classification on labeling for each of the 2 classes.

%==============================table 2=========================== 
\begin{table}[htbp]
\centering
\caption{Predicted results for auto-labeling on the test dataset with P, R and F1 calculated for each of  2 classes}
\label{tab:test}
\begin{tabular}{@{}ccccccc@{}}
\toprule
\multicolumn{2}{c}{\begin{tabular}[c]{@{}c@{}}Confusion\\  matrix\end{tabular}} & Precision & Recall & F1-Score & AUC                    & Accuracy               \\ \midrule
\multicolumn{1}{c|}{487}               & \multicolumn{1}{c|}{13}                & 0.994     & 0.974  & 0.984    & \multirow{2}{*}{0.998} & \multirow{2}{*}{0.984} \\ \cmidrule(r){1-2}
\multicolumn{1}{c|}{3}                 & \multicolumn{1}{c|}{497}               & 0.975     & 0.995  & 0.984    &                        &                        \\ \bottomrule
\end{tabular}
\end{table}
%================================================================

\section{Conclusions}

This paper presents an investigation into performance of a computer vision based classification task on a dataset from the additive manufacturing process. We use a CNN based classifier in combination with transfer learning and active learning strategies. We improved the overall validation accuracy to about 98\%. We also conducted experiments to  investigate the approximate minimum number of labelled samples needed to reach convergence in training. In future work we plan to further investigate the sampling strategies for active learning especially regarding class imbalance problems. We will involve approaches from semi-supervised learning to reinforce the labelling and self training as an extension to the current active learning mechanism.

% \section*{Acknowledgments}

\bibliographystyle{apalike}

\bibliography{bibliography}

\end{document}